\documentclass[11pt]{article}

\usepackage[utf8]{inputenc}
\usepackage[T1]{fontenc}
\usepackage{times}
\usepackage[margin=1in]{geometry}
\usepackage{hyperref}
\usepackage{url}
\usepackage{graphicx}
\usepackage{booktabs}
\usepackage{amsmath}
\usepackage{amssymb}
\usepackage{subcaption}
\usepackage{enumitem}
\usepackage{authblk}
\usepackage[round]{natbib}
\usepackage{titlesec}
\usepackage{fancyhdr}

\fancypagestyle{plain}{\fancyhf{}\lhead{\small Preprint}\rhead{\small\thepage}}

\titleformat{\section}{\normalfont\large\bfseries\scshape}{\thesection}{1em}{}
\titleformat{\subsection}{\normalfont\normalsize\bfseries\scshape}{\thesubsection}{1em}{}

\hypersetup{colorlinks=true, linkcolor=black, citecolor=black, urlcolor=blue}

\title{When Does a Video--Language Model Stop Watching?\\
Reward Strength Controls the Formation and Reversal\\
of Visual Shortcuts in Multimodal RLVR}

\author[]{Zekun Xu}
\affil[]{School of Computer Science, The University of Sydney, Sydney, Australia}
\affil[]{\texttt{xzk20020423@gmail.com}}
\date{}

\newcommand{\onset}{\textsc{onset}}
\newcommand{\pert}{\mathrm{VHS}}
\newcommand{\lam}{\lambda}
\newcommand{\indfun}{\mathbf{1}}

\begin{document}

\maketitle

\begin{abstract}
Reinforcement learning with verifiable rewards (RLVR) is increasingly applied to
large vision--language models (LVLMs), yet outcome-only optimization can drive a
model to stop attending to the video and instead exploit linguistic priors---a
failure we call a \emph{visual shortcut}. While the existence of such
perception bypass is by now documented, \emph{how} it forms, whether it can be
\emph{undone}, and \emph{when} intervention still helps remain open. We treat the
strength of a grounding penalty, $\lam$, as a control knob and characterize the
formation--reversal dynamics of visual shortcuts along the training time axis.
On a held-out, out-of-distribution diagnostic set, we find: (i) a sharp
\emph{onset}---shortcut reliance emerges abruptly over a narrow window of
optimization steps and is robust across random seeds; (ii) a monotone
\emph{dose--response}---increasing $\lam$ progressively suppresses the shortcut,
and at an intermediate dose the trajectory first \emph{forms} and then
\emph{reverses} the shortcut, exposing a hysteresis-like asymmetry between
acquiring and removing it; and (iii) a \emph{critical intervention window}---applying
the penalty before onset arrests shortcut formation, whereas the same penalty
applied after consolidation is markedly less effective. Together these results
recast visual-shortcut collapse not as a binary defect but as a controllable,
time-dependent, and asymmetric process, with direct implications for when and how
strongly to regularize multimodal RLVR.
\end{abstract}

\section{Introduction}
\label{sec:intro}

Reinforcement learning with verifiable rewards (RLVR) has become a standard
recipe for improving the reasoning of large language models, and recent work
extends it to large vision--language models (LVLMs). A recurring hazard, however,
is that optimizing a final-answer reward can teach the model to \emph{ignore the
visual input}: it learns to answer from language priors and surface statistics
rather than from what the video actually shows. We refer to this degenerate
solution as a \emph{visual shortcut}. The phenomenon---discussed under the
umbrella of perception bypass or evaluator exploitation~\citep{reward_hacking_survey}---is
now well documented. Blind-image ablations are striking: replacing informative
frames with blank ones can \emph{improve} accuracy, indicating that the policy has
learned to treat the image as noise~\citep{thinking_deltas}; and VLMs routinely
bypass visual comparison in favor of language priors and semantic
anchors~\citep{vlms_need_words,language_prior_coe,blink}. What is far less
understood is the \emph{process}: visual shortcuts are typically reported as an
endpoint property (``the model ignores vision''), with little characterization of
how the behavior emerges over training, whether it is reversible, and whether the
timing of intervention matters.

This paper studies that process. Our central move is to treat the
\textbf{strength of a grounding penalty}, $\lam$, as an experimental control knob,
and to read out shortcut reliance \emph{densely along the training time axis}
using a perturbation-based diagnostic. Concretely, we measure a
visual-hacking score ($\pert$) that quantifies how invariant a model's answers are
to a visual perturbation that should change the correct answer: a model that
truly watches the video is sensitive to the perturbation, whereas a model relying
on a shortcut is not. By sweeping $\lam$ and tracking $\pert$ through training on a
held-out, out-of-distribution (OOD) set, we can ask formation, reversal, and
timing questions that endpoint evaluations cannot.

\paragraph{Findings.} We report three results.
\textbf{(1) Onset is real and seed-robust (\S\ref{sec:onset}).} On OOD data,
shortcut reliance does not drift in smoothly; it rises abruptly over a narrow
window of optimization steps, and two independent random seeds produce
near-identical onset curves. This identifies a developmental transition rather
than a memorization artifact.
\textbf{(2) Reward strength yields a monotone dose--response with a
formation--reversal asymmetry (\S\ref{sec:dose}).} Increasing $\lam$ monotonically
lowers the plateau level of the shortcut. Strikingly, at an intermediate dose the
trajectory first \emph{forms} a strong shortcut and then \emph{reverses} it later
in training---an asymmetry between acquiring and removing the shortcut that is
reminiscent of hysteresis in driven systems.
\textbf{(3) There is a critical intervention window (\S\ref{sec:intervene}).}
Applying the grounding penalty \emph{before} onset suppresses shortcut formation,
whereas the same penalty applied \emph{after} the shortcut has consolidated is
substantially less able to undo it.
We additionally probe internal representations (\S\ref{sec:repr}) to ask what
changes inside the model as $\lam$ reshapes the shortcut.

\paragraph{Contributions.}
\begin{itemize}[leftmargin=1.4em,itemsep=2pt,topsep=2pt]
\item We introduce a \emph{dose-controlled, time-resolved} study of visual
shortcuts in multimodal RLVR, treating the grounding-penalty strength $\lam$ as a
control knob and reading shortcut reliance densely along training on a held-out OOD
diagnostic.
\item We show that shortcut \onset{} is sharp and \emph{seed-robust}, that the
plateau is a \emph{monotone} function of $\lam$, and that an intermediate dose
\emph{forms and then reverses} the shortcut---an asymmetry between acquisition and
removal.
\item We demonstrate a \emph{critical intervention window}: a grounding penalty
applied before \onset{} prevents the shortcut, whereas the same penalty applied
after consolidation is much less effective.
\end{itemize}
As a secondary, exploratory analysis we also probe internal representations across
the dose axis (\S\ref{sec:repr}); we find a directional, layer-localized tendency
but it falls within bootstrap variability at our sample size, so we report it as a
motivating observation rather than a core claim.

\paragraph{Relation to prior work.} Our contribution is not the observation that
LVLMs can ignore vision, which prior work already documents
\citep{reward_hacking_survey,thinking_deltas,vlms_need_words}. It is the
characterization of the shortcut as a \emph{controllable, time-resolved, and
asymmetric} process using reward strength as the control variable. Studies of
reward-hacking dynamics report onset-like transitions and even non-monotone
rise/retreat/rebound trajectories~\citep{adversarial_reward_auditing,rebound_hacking},
and large-scale empirical studies document that hacking emerges during
optimization across many RL settings~\citep{empirical_hacking_study};
and verifier-grounding can suppress shortcuts~\citep{gaming_verifiers}; but these
are text or code domains, typically at a single reward setting, and without a
\emph{dose} axis that lets one trace formation against reversal. Mechanistic
accounts localize shortcut circuits on single rewards in text-math
settings~\citep{spurious_rewards}. We discuss these distinctions in
\S\ref{sec:related}.

\section{Setup}
\label{sec:setup}

\paragraph{Task and model.} We post-train Qwen3-VL-8B-Instruct, an 8B-parameter
video--language model,
with GRPO-style RLVR \citep{grpo} on a video question-answering objective. The base
verifiable reward combines answer correctness and a light format term,
$r_{\text{base}} = (1-w)\,a + w\,f$, where $a \in \{0,1\}$ is answer correctness,
$f \in \{0,1\}$ checks that the answer is enclosed in the required tag, and
$w{=}0.1$. To study controllable grounding, we add a grounding penalty whose
strength is set by a scalar $\lam \ge 0$:
\begin{equation}
r = r_{\text{base}} - \lam \cdot h, \qquad
h = a \cdot \indfun\!\left[\hat{y}_{\text{pert}} = \hat{y}_{\text{clean}}\right],
\label{eq:penalty}
\end{equation}
where $\hat{y}_{\text{clean}}$ and $\hat{y}_{\text{pert}}$ are the model's answers on
the clean and temporally perturbed video (the same perturbation used by the VHS
diagnostic, \S\ref{app:repro-diag}), and $h$ is a per-sample hacking indicator: it fires
only when the model answers correctly \emph{and} its answer is unchanged under the
perturbation---i.e.\ it got the answer right without using the temporal content. The
penalty thus pushes the policy toward answers that are both correct and genuinely
dependent on the video, and $\lam{=}0$ recovers the pure outcome reward. The penalty
is computed on-policy at every step from the same clean/perturbed rollout pair used
for telemetry, so it adds no separate reward model.

\paragraph{Visual-hacking diagnostic ($\pert$).} To measure shortcut reliance
independently of raw accuracy, we use a perturbation-based score. For each
diagnostic item we generate answers on the clean video and on a perturbed video,
where the perturbation randomly shuffles the temporal order of the video frames,
destroying the temporal structure a faithful viewer would rely on (the individual
frames are unchanged). $\pert$ is the fraction of items on which the answer is
\emph{unchanged} under perturbation; higher $\pert$ indicates stronger reliance on
a visual shortcut (the model ``does not watch''). The diagnostic uses five-way
multiple-choice questions, so a model that ignores the perturbed video and answers
at random would yield $\pert \approx 0.2$ (chance), whereas $\pert$ near $1$
indicates that the answer is essentially invariant to the temporal content. All
diagnostics use greedy decoding.

\paragraph{Held-out OOD evaluation.} Crucially, $\pert$ is measured on a
\emph{held-out, out-of-distribution diagnostic set} that is disjoint from the
training distribution. This ensures that what we call onset reflects an emergent
behavioral change rather than fitting of the training items.

\paragraph{Trajectory fleet.} To separate seed effects from dose effects, we
train a small fleet of trajectories sharing a common origin and varying one factor
at a time (Table~\ref{tab:fleet}). The reading rules are strict: seed robustness
is assessed only by comparing the $\lam{=}0$ runs across seeds; dose response is
assessed only across $\lam \in \{0,1,2\}$ at a fixed seed; intervention timing is
assessed within the intervention runs. We never compare across two factors at once.

\begin{table}[t]
\centering
\caption{Trajectory fleet. All runs share a common origin; each varies a single
factor. $\pert$ is read on the OOD diagnostic set. ``Onset region'' denotes the
step window over which $\pert$ rises sharply for the $\lam{=}0$ runs.}
\label{tab:fleet}
\begin{tabular}{lcccl}
\toprule
Run & $\lam$ & Seed & Steps & Role \\
\midrule
E1 (main) & $0$ & A & 500 & Onset origin; dose baseline \\
sB (seed) & $0$ & B & 80  & Seed-robustness control \\
L1        & $1$ & A & 150 & Intermediate dose (formation then reversal) \\
L2        & $2$ & A & 150 & Strong dose (early suppression) \\
Intervention (pre/mid/post) & $1$ & A & $50$ each & Timing: penalty applied at pre-/at-/post-onset \\
\bottomrule
\end{tabular}
\end{table}

\section{Onset is real and seed-robust}
\label{sec:onset}

\begin{figure}[t]
\centering
\includegraphics[width=0.85\linewidth]{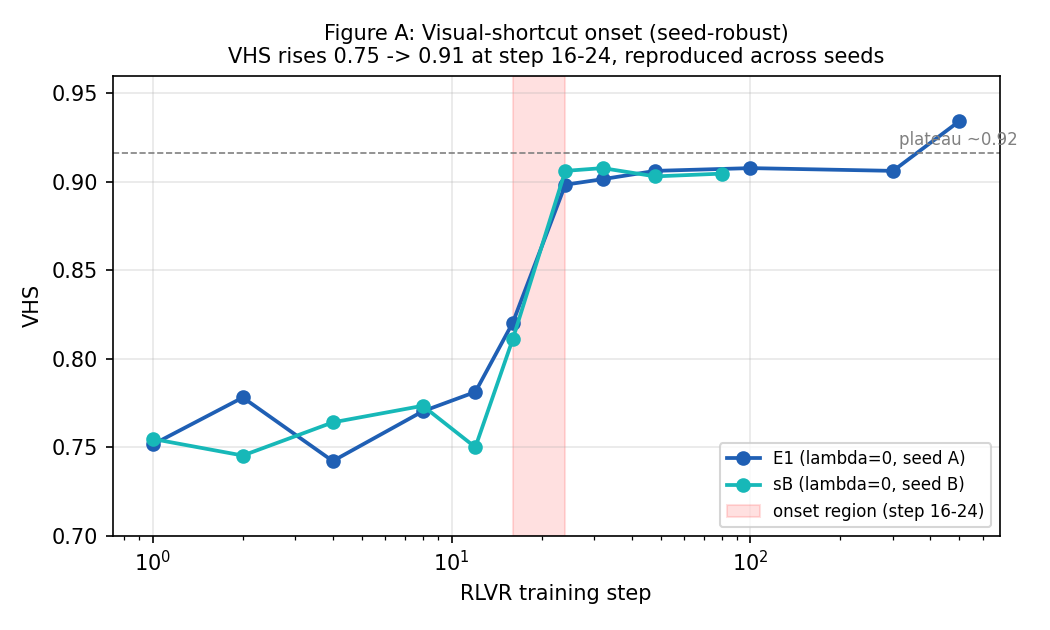}
\caption{\textbf{Onset is real and seed-robust.} On held-out OOD data, two
independent seeds exhibit overlapping, sharply rising $\pert$ curves over the same
narrow step window, indicating an emergent visual-shortcut transition rather than
a memorization artifact.}
\label{fig:onset}
\end{figure}

Figure~\ref{fig:onset} tracks $\pert$ on the OOD diagnostic set as a function of
optimization step for the two $\lam{=}0$ runs (E1, seed~A; sB, seed~B). Three
features stand out. First, the curves are \emph{flat and low} early in training
(through roughly step~16, $\pert \approx 0.75$--$0.78$ with minor step-to-step
fluctuation): the model remains genuinely sensitive to the visual perturbation.
Second, $\pert$ then \emph{rises sharply} over a narrow window---the onset region,
approximately steps~16--24 (shaded in the figure)---reaching a high plateau
($\pert \approx 0.91$) within roughly a dozen steps. Third, and most importantly,
the two seeds produce
\emph{near-identical} onset curves: the rise occurs over the same step window and
saturates at the same level. Because $\pert$ is measured OOD, this reproducibility
indicates that onset is a genuine developmental transition of the policy, not
memorization of training items or a seed-specific accident.

\section{Reward strength: a monotone dose--response with formation--reversal asymmetry}
\label{sec:dose}

\begin{figure}[t]
\centering
\includegraphics[width=0.95\linewidth]{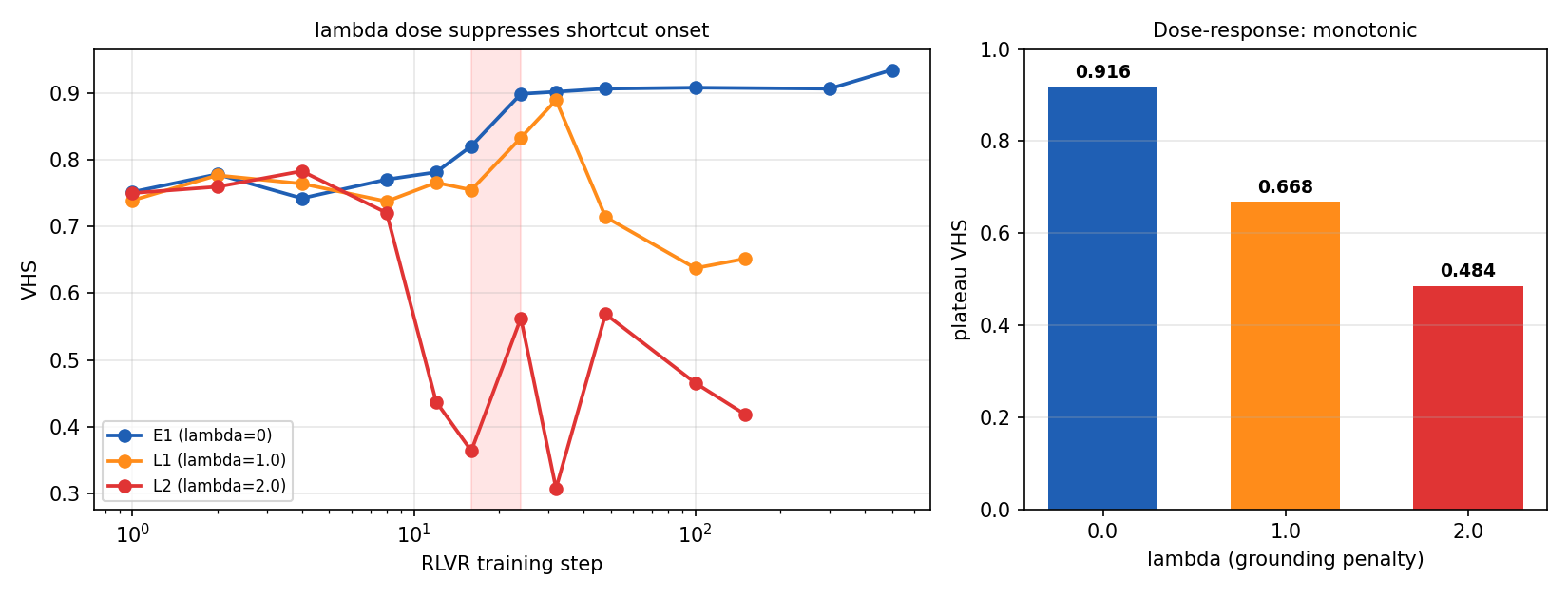}
\caption{\textbf{Dose--response and formation--reversal asymmetry.} Increasing the
grounding penalty $\lam$ monotonically lowers the shortcut plateau (right). At the
intermediate dose, the trajectory forms and then reverses the shortcut (left),
exposing an asymmetry between acquiring and removing it.}
\label{fig:dose}
\end{figure}

We next fix the seed and sweep the grounding penalty $\lam \in \{0,1,2\}$
(runs E1, L1, L2). Figure~\ref{fig:dose} shows both the training dynamics (left)
and the plateau level of $\pert$ as a function of $\lam$ (right).

\paragraph{Monotone dose--response.} The plateau $\pert$ decreases monotonically
with $\lam$: from $\approx 0.91$ at $\lam{=}0$, to $\approx 0.67$ at $\lam{=}1$, to
$\approx 0.48$ at $\lam{=}2$ (Table~\ref{tab:dose}). Each unit of penalty removes a
roughly comparable amount of shortcut reliance, i.e.\ the suppression is
approximately linear in $\lam$ over this range. The seed control (sB, $\lam{=}0$,
plateau $\approx 0.90$) confirms that the $\lam{=}0$ baseline is not seed-specific.

\begin{table}[t]
\centering
\caption{Plateau visual-hacking score $\pert$ on the OOD diagnostic set as a
function of grounding-penalty strength $\lam$ (fixed seed~A; sB is the $\lam{=}0$
seed control). Higher $\pert$ means stronger reliance on a visual shortcut.
$\Delta$ is the change from the $\lam{=}0$ baseline. The intermediate dose
additionally exhibits a non-monotone \emph{form-then-reverse} trajectory in time
(peak before plateau), reported in the last column.}
\label{tab:dose}
\begin{tabular}{lcccc}
\toprule
Run & $\lam$ & Plateau $\pert$ & $\Delta$ vs.\ $\lam{=}0$ & In-time trajectory \\
\midrule
E1        & $0$ & $0.91$ & ---     & rise to plateau \\
sB (seed) & $0$ & $0.90$ & $-0.01$ & rise to plateau \\
L1        & $1$ & $0.67$ & $-0.24$ & \textbf{form then reverse} (peak $\approx0.89$) \\
L2        & $2$ & $0.48$ & $-0.43$ & early suppression, larger fluctuation \\
\bottomrule
\end{tabular}
\end{table}

\paragraph{Formation--reversal asymmetry.} The intermediate dose is the most
informative. At $\lam{=}1$ (L1), $\pert$ does not simply settle at a lower level:
it first \emph{rises} to a peak comparable to the unpenalized run (forming a strong
shortcut), and only later \emph{falls} as training continues (reversing it). The
strong dose $\lam{=}2$ (L2) instead suppresses the shortcut earlier and exhibits
larger fluctuations, consistent with a stronger but less stable drive. The key
qualitative point is that \emph{forming} the shortcut and \emph{removing} it are
not mirror images along the time axis. We describe this loosely as
hysteresis-like, but we use the term only by analogy: we observe a non-monotone
form-then-reverse trajectory in time at a fixed dose, not a true hysteresis loop
traced by sweeping $\lam$ up and then down. Establishing genuine path dependence
would require such a bidirectional dose sweep, which we leave to future work; here
the asymmetry refers specifically to the difference between how quickly the shortcut
forms and how slowly it reverses within a single run.

\paragraph{Accuracy corroborates ``watching vs.\ guessing.''} On the same OOD set,
raw accuracy tracks $\pert$ across the dose axis: at plateau the unpenalized run
($\lam{=}0$) reaches $\pert{\approx}0.91$ with accuracy $0.82$; at $\lam{=}1$, $\pert$
falls to $0.67$ and accuracy to $0.61$; and at $\lam{=}2$, $\pert{\approx}0.48$ with
accuracy $0.55$ (all averaged over the plateau checkpoints). The per-sample hacking
rate $h$ from Eq.~\ref{eq:penalty}---answered correctly \emph{and} invariant to the
perturbation---falls in lockstep, from $0.78$ at $\lam{=}0$ to $0.47$ and $0.38$.
The unpenalized model thus answers confidently without watching (high $\pert$, high
accuracy, high $h$), and as the penalty suppresses the shortcut, accuracy declines
by far less than $\pert$ does (a $0.43$ drop in $\pert$ versus a $0.27$ drop in
accuracy from $\lam{=}0$ to $\lam{=}2$). This gap is consistent with the
interpretation that the shortcut inflates apparent accuracy, and that suppressing it
removes the inflated portion while leaving a substantial vision-grounded core, rather
than uniformly degrading a genuine capability.

\section{A critical intervention window}
\label{sec:intervene}

If forming and reversing a shortcut are asymmetric, the \emph{timing} of
intervention should matter. We test this directly. Starting from checkpoints of
the unpenalized run E1 taken before onset, at the onset transition, and after
consolidation, we resume training with the grounding penalty ($\lam{=}1$) for a
fixed number of additional steps in each case (the pre-/at-/post-onset
intervention runs), holding the number of intervention steps constant so that
differences cannot be attributed to training duration.

\begin{figure}[t]
\centering
\includegraphics[width=0.95\linewidth]{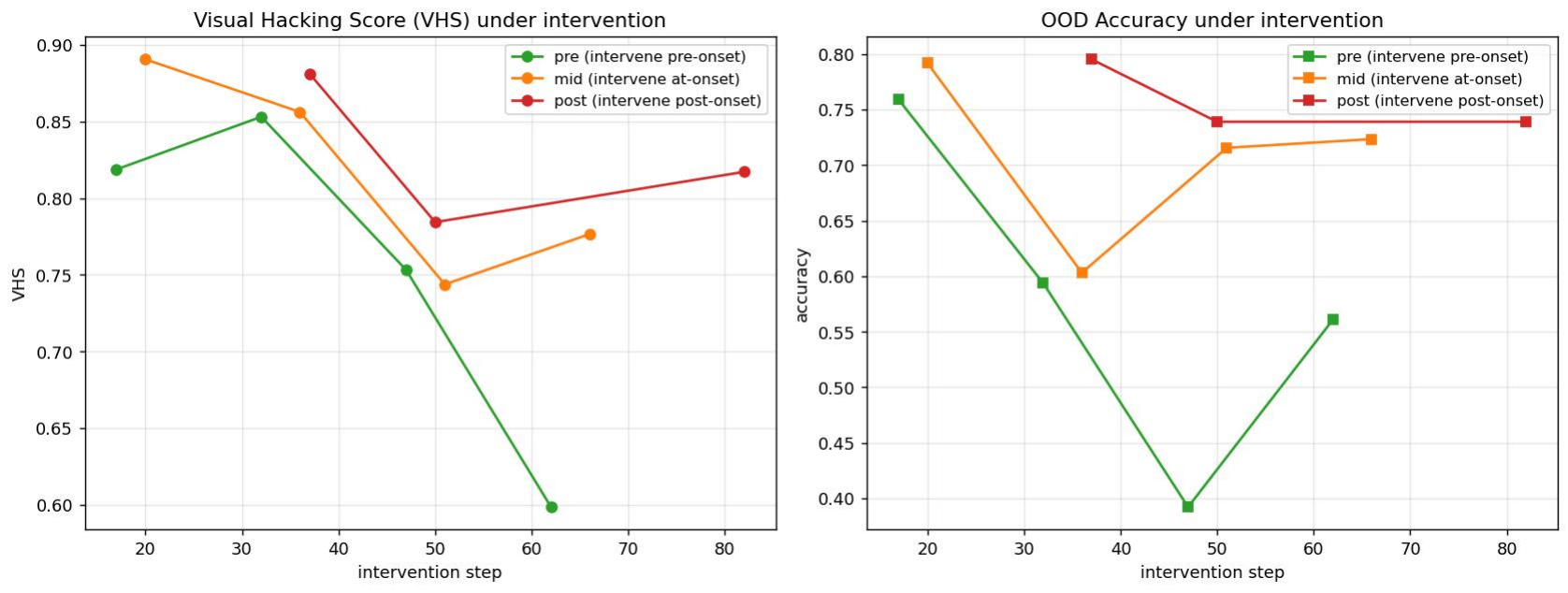}
\caption{\textbf{A critical intervention window.} The same grounding penalty
suppresses the shortcut when applied before onset but is much less able to reverse
it once the shortcut has consolidated, demonstrating time-dependent
intervenability.}
\label{fig:intervene}
\end{figure}

Figure~\ref{fig:intervene} reports $\pert$ during each intervention. Intervening
\emph{before} onset drives $\pert$ down over the intervention budget, reaching
$0.60$ by the end---the penalty prevents the shortcut from consolidating.
Intervening \emph{after} consolidation is markedly less effective: $\pert$ remains
elevated (around $0.78$--$0.88$ throughout), indicating that once the shortcut has
formed it is comparatively resistant to the same corrective pressure. The at-onset
case is intermediate, declining only partway (to roughly $0.74$--$0.78$). This is
the actionable consequence of the asymmetry in \S\ref{sec:dose}: there is a
window---around and before onset---in which a modest grounding penalty is most
effective, and waiting until the behavior has consolidated forfeits much of that
leverage.

\section{What changes inside: representation probe}
\label{sec:repr}

To ask what $\lam$ reshapes internally, we conduct a \emph{preliminary} probe of
hidden-state activations from the language-model decoder layers of matched
checkpoints across doses ($\lam \in \{0,1,2\}$) at fixed steps, on a stratified
diagnostic subset, summarizing their geometry with unsupervised metrics (effective
rank, anisotropy, and mean norm). The pattern we observe is \emph{layer-localized}:
early layers (which encode low-level perceptual features) and the final layers
(near the output) are essentially unchanged across doses, whereas the middle
layers---where semantic and reasoning content is typically represented---show a
dose-dependent tendency, with the largest cross-$\lam$ spread concentrated in the
middle of the network and a parallel pattern in anisotropy. We emphasize that this
is an exploratory observation: at the present sample size the mid-layer spread is
suggestive but lies within bootstrap variability, so we report it as a directional
finding rather than a calibrated effect, and a higher-powered analysis is left to
future revision. Taken together with the behavioral results, it offers a tentative
localization---grounding pressure appears to act on mid-network geometry rather
than rescaling the representation uniformly---that motivates a more careful
mechanistic study.

\begin{figure}[t]
\centering
\includegraphics[width=0.95\linewidth]{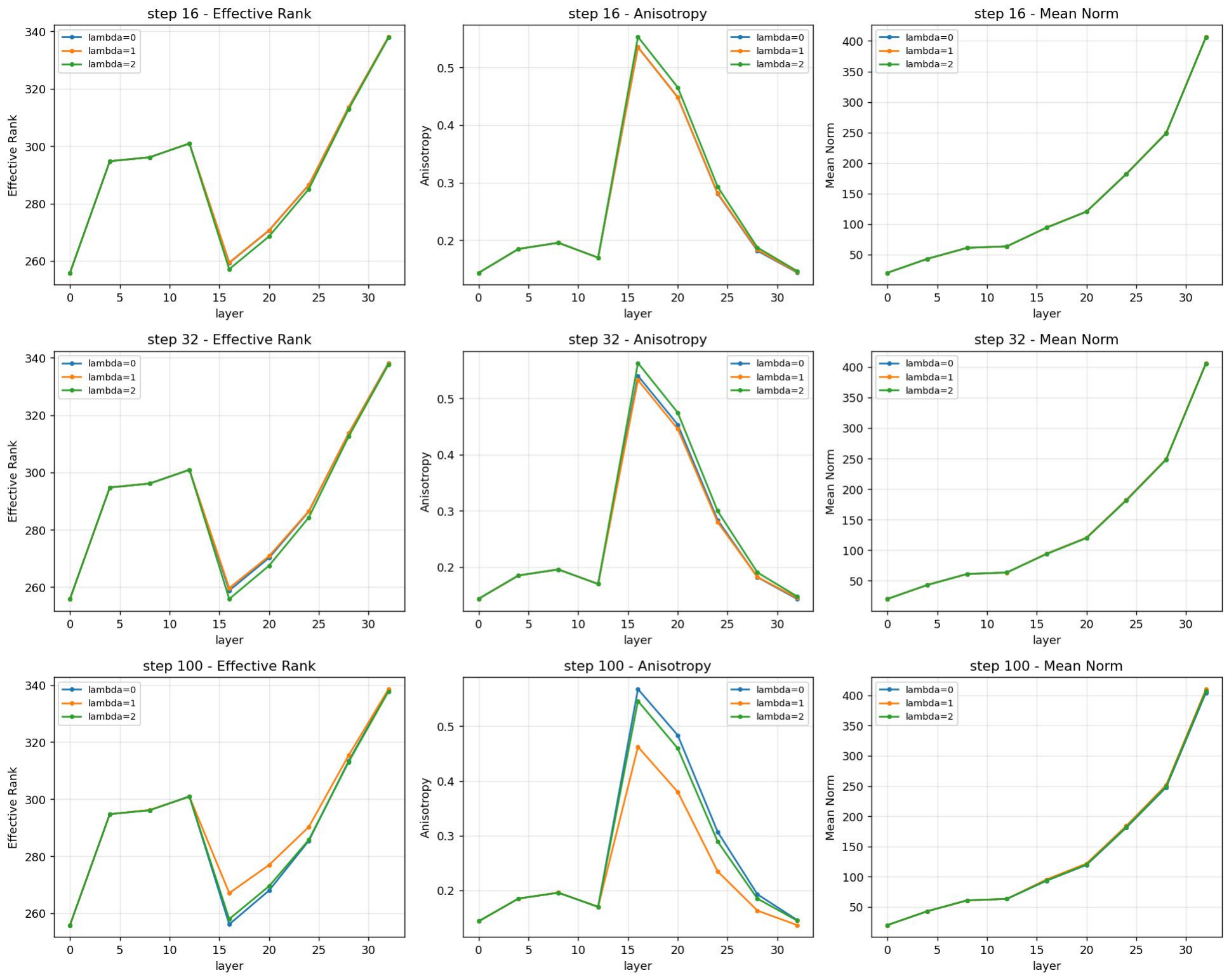}
\caption{\textbf{A preliminary, layer-localized representation signature.} Internal
representation statistics of matched checkpoints, contrasting $\lam{=}0,1,2$. The
observed dose-dependent tendency is concentrated in the middle layers and near-zero
at early and final layers; we report it as an exploratory, directional finding (the
mid-layer spread is within bootstrap variability at the current sample size) that
motivates a more careful mechanistic study.}
\label{fig:repr}
\end{figure}

\section{Related work}
\label{sec:related}

\paragraph{Perception bypass in (multimodal) RLVR.}
A growing body of work documents that LVLMs under-use visual evidence and rely on
language priors. Surveys frame this as perception bypass / evaluator
exploitation~\citep{reward_hacking_survey}; blind-image ablations show accuracy can
rise when the image is removed~\citep{thinking_deltas}; and controlled benchmarks
isolate textual-prior reliance~\citep{vlms_need_words,language_prior_coe,see_or_guess,blink}.
These establish that the shortcut \emph{exists}. We instead use reward strength as
a control variable to characterize \emph{how} it forms, whether it reverses, and
\emph{when} intervention helps.

\paragraph{Temporal dynamics and onset of reward hacking.}
Several works observe that reward hacking is not present from the start but emerges
at a characteristic phase of optimization: large-scale empirical studies across many
RL environments and algorithms catalogue reward-hacking categories and find that
hacking arises during training rather than being present from
the outset~\citep{empirical_hacking_study}; auditing frameworks treat hacking as
a dynamic, detectable signal that arises during
optimization~\citep{adversarial_reward_auditing}; and dedicated testbeds study the
emergence and generalization of hacking~\citep{countdown_code}. Most relevantly,
rebound hacking exhibits a non-monotone fail/retreat/rebound trajectory that is
robust across seeds~\citep{rebound_hacking}. Our intermediate-dose run shows a
related form-then-reverse pattern, but we obtain it by \emph{tuning reward
strength} and read it out with an OOD visual diagnostic, yielding a dose-resolved
account rather than a single-setting observation.

\paragraph{Suppression, verifiers, and mechanisms.}
Grounding the verifier can remove shortcuts: e.g.\ isomorphic (rather than merely
extensional) verification eliminates rule-induction shortcuts~\citep{gaming_verifiers},
echoing our use of a grounding penalty---though as a binary design choice rather
than a continuous dose. Mechanistic analyses localize shortcut circuits and show
bidirectional steering on single rewards in text-math settings~\citep{spurious_rewards}.
Our representation probe is complementary, contrasting matched checkpoints across a
reward-strength axis and along the training-time axis in a multimodal setting.

\section{Discussion and limitations}
\label{sec:discussion}

We have shown that visual-shortcut collapse in multimodal RLVR is better described
as a controllable, time-dependent, and asymmetric process than as a binary defect.
Reward strength acts as a knob that sets the plateau level of shortcut reliance;
the formation and reversal of the shortcut are asymmetric in time; and there is a
critical window---around and before onset---in which a modest grounding penalty is
most effective. Practically, this argues for grounding regularization that is
\emph{early} and \emph{dose-calibrated} rather than applied late as a remedy.

\paragraph{Limitations.} Our results should be read as a detailed case study of one
video--language model rather than a universal law of RLVR. All trajectories use a
single model family (Qwen3-VL-8B-Instruct) and a single grounding-penalty design, so
we cannot say whether the specific onset location, its sharpness, or the
near-linear dose response transfer to other scales (e.g.\ 2B or 30B+ variants) or to
other architectures (e.g.\ LLaVA- or InternVL-style models); the existence of an
abrupt onset and of a dose- and timing-dependent intervention effect is what we
claim, not its precise quantitative form. Establishing how the onset step and dose
sensitivity scale with model size and shift across architectures is the most
important direction for future work, and the diagnostic and penalty defined here are
designed to port directly to such a study. The $\pert$ diagnostic also
measures invariance to a particular class of visual perturbations (temporal frame
shuffling) and is a proxy
for ``watching''; alternative diagnostics may reveal complementary facets. Finally,
the representation analysis is correlational and, as noted in \S\ref{sec:repr},
falls within bootstrap variability at our sample size; it is intended to relate, not
to causally explain, behavioral shortcut suppression.

\bibliographystyle{plainnat}
\bibliography{vgonset}

\appendix
\section{Reproducibility and diagnostic details}
\label{app:repro}

This appendix records the diagnostic construction, the trajectory fleet, the
evaluation protocol, and the representation-extraction pipeline in enough detail to
reproduce the trends reported in the main text.

\subsection{Model and training}
The base policy is an 8B-parameter video--language model (Qwen3-VL-8B-Instruct;
36 decoder layers, hidden size 4096). We post-train with GRPO-style
RLVR~\citep{grpo} on a video question-answering objective. The base verifiable
reward is answer correctness; the grounding penalty is added with strength
$\lam \ge 0$, where $\lam{=}0$ recovers the pure outcome reward. During RL we use a
global batch size of $512$, freeze the vision tower, and keep the visual
input pipeline fixed across runs so that differences between trajectories are
attributable to the reward configuration rather than to data ordering or
preprocessing.

\subsection{Trajectory fleet}
All runs share a common origin and vary a single factor (Table~\ref{tab:fleet}):
E1 ($\lam{=}0$, seed~A, 500 steps) is the onset origin and dose baseline; sB
($\lam{=}0$, seed~B, 80 steps) is the seed-robustness control; L1 ($\lam{=}1$) and
L2 ($\lam{=}2$) are the intermediate and strong doses (150 steps each); the
intervention runs (pre-/at-/post-onset) resume from E1 checkpoints taken before
onset, at the onset transition, and after consolidation, each trained for a fixed
50 additional steps with $\lam{=}1$. The reading rules are strict: seed robustness
is assessed only across the $\lam{=}0$ runs; dose response only across
$\lam \in \{0,1,2\}$ at fixed seed; intervention timing only within the
intervention runs.

\subsection{Held-out OOD diagnostic and the VHS perturbation}
\label{app:repro-diag}
The visual-hacking score (VHS) is read on a held-out, out-of-distribution
diagnostic set of $640$ five-way multiple-choice questions built from the NExT-QA
video-QA distribution, disjoint from the RLVR training split, so that the measured
onset reflects an emergent behavioral change rather than fitting of the training
items. The questions are temporally grounded (they ask about the order or
progression of actions and events). For each item we generate an answer on the
clean video and on a perturbed video. The perturbation randomly permutes the
temporal order of the sampled frames (a random shuffle of the frame sequence),
which destroys the temporal structure---the ordering of actions and events---that
a temporally grounded question depends on, while leaving the individual frames and
their appearance statistics intact. A model that genuinely attends to the video
should therefore change its answer under this perturbation, whereas a model
answering from language priors or static cues is unaffected. VHS is the fraction of
items whose answer is \emph{unchanged} under the perturbation: higher VHS indicates
stronger reliance on a visual (specifically temporal) shortcut. Because the
questions are five-way, a model that ignores the perturbed video and answers at
random gives VHS $\approx 0.2$ (chance), while VHS near $1$ means the answer is
essentially invariant to the temporal content. We also record clean accuracy and
the hacking rate on the same set as corroborating readouts. (This behavioral
diagnostic is distinct from the data used for the representation probe in
\S\ref{app:repr-extract}, which draws on a separate video-understanding benchmark
purely as a source of inputs for activation extraction.)

\subsection{Evaluation protocol}
All diagnostics use greedy decoding (temperature $0$, a single sample per item).
Each checkpoint is evaluated by loading the trained actor and running the
clean/perturbed double generation on the diagnostic set with a fixed batch size,
maximum prompt length $8192$, maximum response length $1024$, and a fixed video
sampling rate; the visual token budget is held constant across checkpoints. To
build a trajectory we evaluate a sequence of saved checkpoints under identical
settings and plot the resulting VHS against the optimization step. Evaluation
configuration (decoding, batching, and the diagnostic set) is held fixed across all
runs so that trajectories are directly comparable; old and new evaluation
configurations are never mixed within a single figure or table.

\subsection{Representation-extraction pipeline}
\label{app:repr-extract}
For the representation probe we extract hidden-state activations from the
language-model decoder layers of matched checkpoints. We read a fixed set of layers
(every fourth decoder layer, i.e.\ layers $0,4,8,\dots,32$), mean-pool token
activations within each item, and accumulate a fixed, stratified set of input
items per checkpoint, sampled from a standard video-understanding
benchmark~\citep{mvbench}. We emphasize that this probe uses MVBench purely as a
fixed pool of video inputs for activation extraction; it is a separate data source
from the NExT-QA--based VHS diagnostic of \S\ref{app:repro-diag} and is not used to
compute VHS or any behavioral metric. For each (checkpoint, layer) we summarize the
resulting activation matrix with three unsupervised geometry metrics: effective
rank (the exponential of the entropy of the normalized singular-value spectrum),
anisotropy (the share of variance captured by the top principal direction), and
mean activation norm. The dose-dependent spread reported in the main text is the
range of a metric across $\lam \in \{0,1,2\}$ at a matched checkpoint and layer. As
noted in \S\ref{sec:repr}, at the sample size used here the mid-layer spread is
suggestive but lies within bootstrap variability; we therefore report it as a
directional, exploratory finding and leave a higher-powered analysis to future
revision.

\end{document}